\documentclass[runningheads]{llncs}
\usepackage{graphicx}
\newcommand{\dataname}{\texttt{3Cext}}
\newcommand{\modelname}{\texttt{PeriCite}}
\usepackage{amsmath}

\usepackage{amssymb}

\usepackage{booktabs, multirow} 
\usepackage{soul}
\usepackage[table]{xcolor} 
\usepackage{todonotes}

\begin{document}
\title{Inline Citation Classification using Peripheral Context and Time-evolving Augmentation}
%
%
\author{
	Priyanshi Gupta$^1$,
    Yash Kumar Atri$^1$,
	Apurva Nagvenkar$^2$, \\ Sourish Dasgupta$^2$, Tanmoy Chakraborty$^3$ }
 \institute{$^1$IIIT-Delhi, India, $^2$Crimson AI, India, 
        $^3$IIT Delhi, India \\
	{\tt  \{priyanshig,yashk\}@iiitd.ac.in, \{apurva.nagvenkar,sourish.dasgupta\}@crimsoni.ai, \\  tanchak@iitd.ac.in}
}

%
%
%
\maketitle              
\begin{abstract}
Citation plays a pivotal role in determining the associations among research articles. It portrays essential information in indicative, supportive, or contrastive studies. The task of inline citation classification aids in extrapolating these relationships; However, existing studies are still immature and demand further scrutiny. Current datasets and methods used for inline citation classification only use citation-marked sentences constraining the model to turn a blind eye to domain knowledge and neighboring contextual sentences. In this paper, we propose a new dataset, named \dataname, which along with the cited sentences, provides discourse information using the vicinal sentences to analyze the contrasting and entailing relationships as well as domain information. We propose {\modelname}, a Transformer-based deep neural network that fuses peripheral sentences and domain knowledge. Our model achieves the state-of-the-art on the {\dataname} dataset by $+0.09$ F1 against the best baseline. We conduct extensive ablations to analyze the efficacy of the proposed dataset and model fusion methods.

\end{abstract}
\section{Introduction}
For the past several decades, there has been an interest in citation analysis for research evaluation. Researchers have emphasized the necessity for new methodologies that take into account various components of citing sentences. A well-known qualitative technique for assessing the scientific influence is to analyze the sentence in which the research article is mentioned to ascertain the purpose behind the citation. The context of the citation, or the text in which the cited document is mentioned, has proven to be an effective indicator of the citation's intent \cite{teufel2006automatic}. Measuring the scientific impact of research articles requires a fundamental understanding of citation intent. A great way to gauge the significance of a scientific publication is to determine why citations are made in one's work and how significant they are. 

Previous methods for citation context categorization used a range of annotation techniques with low-to-high granularity. Comparing the earlier systems is extremely difficult due to the absence of standardized methodologies and annotation schemes. The 3C shared task \cite{kunnath2020overview,kunnath2021overview} used a piece of the Academic Citation Typing (ACT) dataset to categorize the reference anchor into `function' or `purpose' by looking at the citing sentence or the text that contains the citation \cite{pride2019act}. Only quantitative elements are considered in traditional citation analysis based solely on the citation count. One of the biggest obstacles to citation context analysis for citation identification is that there is no multidisciplinary dataset and that there isn't any medium to fine-grained schemes that adequately represent the function and its influence \cite{hernandez2015citation}. To address this challenge, Kunnath et al. \cite{kunnath2020overview} provided a unified task, called the 3C Shared Task, to compare several citation classification approaches on the same dataset to address the shortcomings of citation context categorization. The main distinction in the second iteration of this task \cite{kunnath2021overview} was that the subtasks contained full-text datasets. However, even with the full text, the metadata associated with the citation sentence was not adequate to understand the reasoning for the citation.

To alleviate the above limitations, we propose a new dataset, named \dataname, and a new model, named \modelname\ that combines the advantages of augmentation and peripheral context. Experiments show that the cited sentences heavily rely on the peripheral context to strengthen an argument by contrasting or entailing information. 
Our main contributions are as follows
\begin{enumerate}
    \item We extend the 3C dataset \cite{kunnath2021overview} -- \dataname, which, along with the cited sentence, adds more discourse information by providing contrasting and entailing information using the peripheral sentences.
    \item We propose a novel model, \modelname, which uses spatial fusion and cross-text attention to attend to contextual information for the peripheral sentences and time-evolving augmentation to counter class imbalance during the training time.
    \item We also compare our proposed model against various baselines and show the efficacy of the module along with ablation studies and error analysis.
\end{enumerate}

\section{Related Work}

Citations are important for persuasion since they provide a source of support for the assertions made by authors. Understanding whether the writers agree or disagree with the assertions made in the cited publication is crucial because not all citations are used with a similar purpose. In order to classify citations according to their context, a sizable corpus of research has previously examined the language used in scientific discourse. Several frameworks have been devised to categorize the intent of citations \cite{moravcsik1975some}. Many strategies were used in the early efforts for automatic citation intent categorization; they included rule-based systems \cite{garzone2000towards,pham2003new}, machine learning techniques based on language patterns \cite{RePEc:spr:scient:v:119:y:2019:i:1:d:10.1007_s11192-019-03028-9}, and manually-constructed features from the citation context. Teufel et al. \cite{teufel2006automatic} introduced how to annotate citations in scholarly articles for 12 classes and used machine learning techniques to replicate annotation. These classes were split into four top-level groups, namely neural class, citations that expressly address weaknesses, citations that contrast or compare, and citations that concur with, use, or are compatible with the citing work. Abu-Jbara et al. \cite{abu2013purpose} utilized a linear SVM and lexical, structural, and syntactic characteristics for categorization. Additionally, feature-based techniques \cite{cohan2015matching,cohan2017contextualizing} for locating quoted spans in the mentioned publications have been studied. Improvements were shown by a joint prediction of cited spans and citation function using a CNN-based model \cite{su2019neural}.

 Most of these initiatives offer far too fine-grained citation categories, some of which are infrequently used in articles. They, therefore, serve little purpose in automated analyses of scientific articles. Jurgens et al. \cite{jurgens2018measuring} developed a six-category technique to incorporate earlier research and suggested a more precise categorization scheme expanding all previous feature-based work on citation intent classification. The authors also added six categories and $1,941$ samples from computational linguistics studies in addition to the three original features—pattern-based, topic-based, and prototype argument-based. They also used structural topology, lexical semantics, grammatical, field positions and values, and usage characteristics.
 
All of the methods listed above, classified data using hand-engineered features. Cohan et al. \cite{cohan2019structural}  proposed a neural multi-task learning technique for classifying citation intent using non-contextualized (GloVe \cite{pennington-etal-2014-glove}) and contextualized embeddings (ELMo \cite{sarzynska2021detecting}, Bidirectional LSTM, and attention method). The authors used two auxiliary tasks to support the primary classification task in order to accomplish multi-task learning. Their recent research  \cite{cohan2019structural} included only three citation categories and $11,020$ instances from the Computer Science and Medical domains to make up their new dataset (SciCite). Beltagy et al. \cite{beltagy2019scibert} released SciBERT, a model pre-trained over $1.14$ million papers from Semantic Scholar. To support the study in this area, a recent analysis by \cite{kunnath2022meta} evaluated 60 research articles on this topic, the difficulties the researchers had while conducting their work, and the knowledge gaps that still need to be addressed.


\section{Methodology}
\begin{figure*}[t]
    \centering
    \scalebox{0.97}{
    \includegraphics[width=\textwidth]{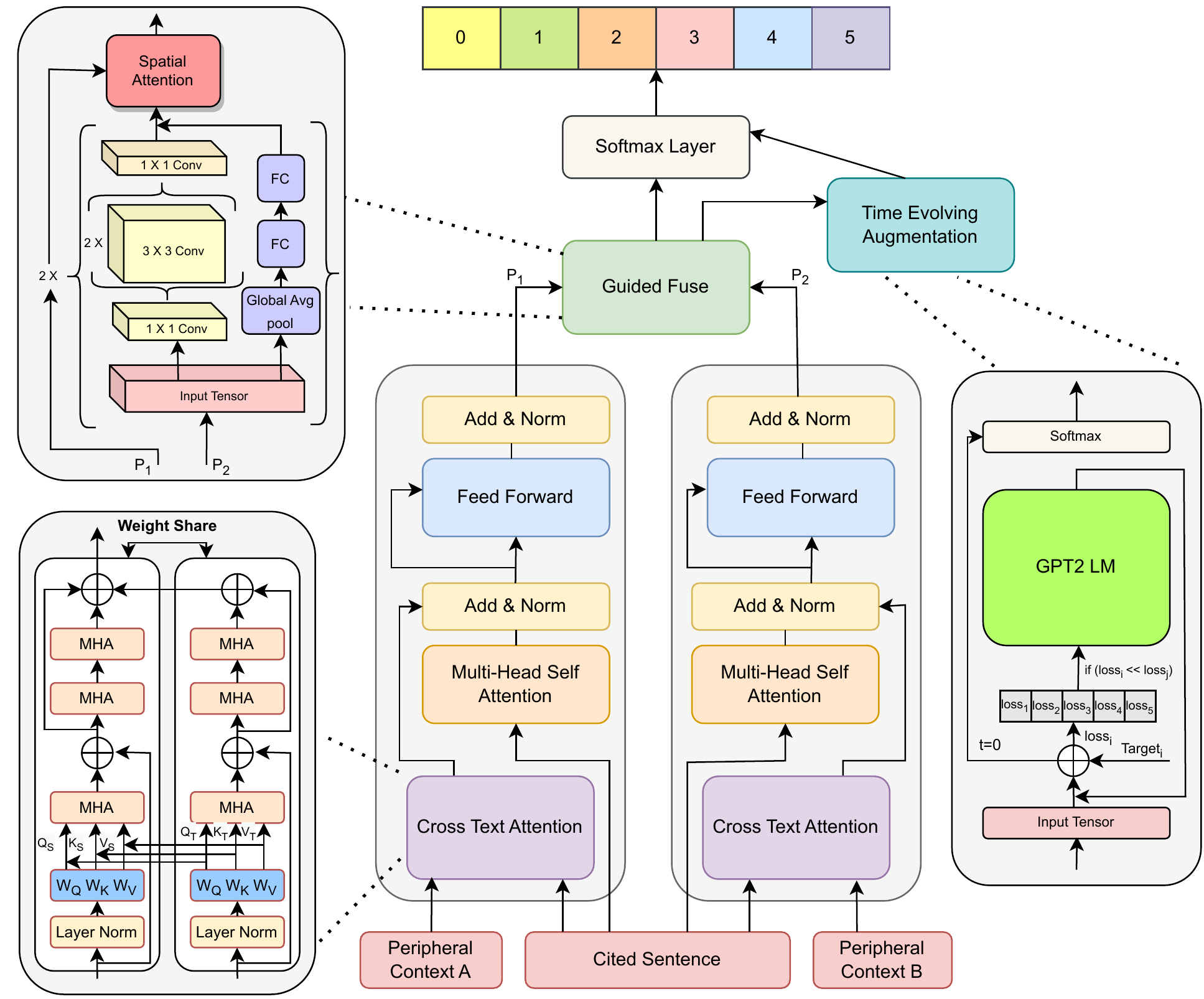}}
    \caption{{\bf Illustration of \modelname}: It comprises two parallel stacked Transformer blocks. The cross-text attention is computed between the main context and peripheral sentence. The output from the Transformer blocks is fused using the Guided Fuse. The time-evolving augmentation based on the label representation in the mini-batch generates synthetic training samples.}
    \label{fig:model}
    \vspace{-5mm}
\end{figure*}

In this section, we discuss our proposed model, \modelname. It comprises two stacked Transformers, each with four blocks. It uses Cross-Text Attention to capture the discourse between the cited and peripheral sentences. \modelname\ also houses Time-evolving Augmentation to synthetically generate data as per label loss and Spatial Fusion to fuse the final representations of stacked Transformers. Figure \ref{fig:model} shows the schematic diagram of \modelname. 

\subsection{Cross-Text Attention}
Attention formulation over a single input text may not provide adequate information to the model. However, when fused with the peripheral context attention, the model can learn important excerpts relative to the label. To fuse peripheral context attention to the main input, we propose Cross-Text Attention (CTA). It computes pairwise weights between the main text and a peripheral context.

Given the self-attention as
\begin{equation} \label{eq:atten_self}
    Attn_{\text{self}} (\mathbf{x}) = softmax \left(\frac{QK^{T}}{\sqrt{d_{k}}}\right) V,  \nonumber
\end{equation}
 initially, queries \( Q \in \mathbb{R}^{N \times d_{k}} \) , keys \begin{math} K \in \mathbb{R}^{N \times d_{k}} \end{math}, and values \( V \in \mathbb{R}^{N \times d_{v}} \) are generated for the main input text with $d_k$ and $d_v$ as their dimensions, respectively. Next, to compute the CTA score, pairwise weights between the main input and peripheral context are computed using 

\begin{equation}
        Attn_{bidir} (x_s, x_t) = Attn_{cross} (x_t, x_s) + Attn_{cross} (x_s, x_t), \nonumber
\end{equation}        
\begin{equation}
        Attn_{cross} (x_t, x_s) = softmax \left(\frac{Q_t K_s^{T}}{\sqrt{d_{k}}}\right) V_s,  \nonumber
\end{equation}        
\begin{equation}
Attn_{cross} (x_s, x_t) = softmax \left(\frac{Q_s K_t^{T}}{\sqrt{d_{k}}}\right) V_t\nonumber
\end{equation}
Here $x_s$ denotes the contextual representation of the main input text, $x_t$ denotes the context of the peripheral text, and $Q$, $K$, and $V$ with $s$ and $t$ represents queries, keys, and values based on the main and peripheral text, respectively. We then perform a linear projection of attention heads to capture language comprehension. Finally, the computed attention weights are passed through a feed-forward layer.

\subsection{Spatial Fusion}

Since the fusion module combines the generated attention vectors from both peripheral encoders, it is crucial to determine which vector is more significant, how it contributes to the main context and interacts with each other. Keeping this intuition in mind, we utilise a fusion strategy based on Spatial Fusion (SF). SF was first introduced by Li et al. \cite{li2020nestfuse} to fuse multiple images to decipher deep features. We extend it in our setting to fuse texts together to form deep contextual features.

In spatial fusion as suggested in \cite{vs2022image}, global features are modeled by a self-attention layer, which is obtained from a convolutional and a dimensionality reduction layer. The combination of these two captures the local as well as the global relations between the feature set. Further, we pass the obtained features to a range of convolutional blocks to enhance the contexts during fusion. Finally, an ordered weighting average map is created to merge features with the source text. 

The multi-scale feature matrix is denoted by $a$, where $a$ is the list of tensors $a\in{\{1,2,\cdots,a\}},a=4$.
The fusion mechanism takes two inputs -- one from the cited sentence as $P_1^a$, and the other from the peripheral sentence as $P_2^a$. The spatial attention \cite{li2020nestfuse} $C_f^a$ is created by the fusion and fed to the final decoding layer.

The Ordered Weighing maps are computed using the L1 normalization and softmax over $P_1^a$ and $P_2^a$ resulting in weight matrix $W_1^a$ and $W_2^a$, respectively. The final Ordered Weighing maps are computed using Eq. \ref{Eq5} as follows,

\begin{eqnarray}\label{Eq5}
  	S_j^a(m,n) = \frac{P_j^a(m,n)}{\sum_{i=1}^j||P_i^a(m,n)}
\end{eqnarray}
Here L1 norm is computed for both $P_j^a$ and $P_i^a$ with $j$ ranging in $[0,2]$ set. The position in the feature set $P_j^a$ and $P_i^a$ are indicated by ($m,n$) with a fixed vector size $V$. The $P_j^a(m,n)$ also outputs a vector of size V.

The feature vectors $P_1^a$ and $P_2^a$ over the weight matrix $W_1^a$ and $W_2^a$ are further enhanced by weighing them using $\alpha_k^m$ using Eq. \ref{Eq6},	
\begin{eqnarray}\label{Eq6}
  	P_j^a(m,n) = \alpha_j^a(m,n) \times P_j^a(m,n)
\end{eqnarray}

Finally, the fused vector $C_f^a$ is computed by projecting it against the enhanced feature set using Eq. \ref{Eq7}

\begin{eqnarray}\label{Eq7}
  	C_f^a(m,n) = \sum_{i=1}^j P_j^a(m,n)
\end{eqnarray}

\subsection{Time Evolving Augmentation}

Data augmentation is useful to generate synthetic data and to balance a dataset. However, most of the data augmentation techniques generate synthetic data based on random transformations of the minor class. As shown in Table \ref{tab:class_count}, our proposed dataset \dataname\ also shows heavy class imbalance with the majority class showing three times the number of samples than the second major class. Two of the most popular ways to handle the class imbalance is either to make all classes  equally representative in the training set or augment the minority class's samples to match the majority class. However, for the task of inline citation classification, both of these methods lead to more degraded performance pertaining to structural complexity and information spread. To tackle these limitations, we propose Time-evolving Augmentation (TEA).

At every time step $t$, TEA computes the label representation in each mini-batch $m$ as $s_i = [l_1,l_2,l_3,l_4,l_5]$. For a loss computed at time step $t$, the model computes the loss per label $l_i$ for a given mini batch $m$ and formulates a loss to label relationship $loss \rightarrow label$ as $losslabel_i$. Given the distribution $losslabel_i$, TEA synthetically generates training samples for the minority class having the highest loss in a given mini-batch. The data samples are augmented using the GPT2 language model \cite{radford2019language}. The $loss \rightarrow label$ representation is independent of the global representation of a number of samples per class and only takes into account the representation of the given mini-batch. This method helps the model keep the loss in check for each label at every step. The loss of representation per label is a guiding factor for the TEA to evolve at every time step, helping to model to learn equal representation during the training phase.

\section{Experiments}

\begin{table}[!t]\centering

\scriptsize
\begin{tabular}{p{0.3\linewidth}|p{0.3\linewidth}|p{0.3\linewidth}}\toprule
First Sentence &Cited Sentence &Second Sentence \\\cmidrule{1-3}
[CITE] describe a hybrid recommender system that exploits ontologies to increase the accuracy of the profiling process and hence the usefulness of the recommendations. &\#AUTHOR\_TAG use a different strategy by representing user profiles as bags-of-words and weighing each term according to the user interests derived from a domain ontology &Razmerita et al. [CITE] describe OntobUM, an ontology-based recommender that integrates three ontologies: i) the user ontology, which structures the characteristics of users and their relationships, ii) the domain ontology, which defines the domain \\\cmidrule{1-3}
Content-based recommender systems [CITE] rely on pre-existing domain knowledge to suggest items more similar to the ones that the user seems to like. They usually generate user models that describe user interests according to features [CITE]. & This API supports a number of applications, including Smart Book Recommender, Smart Topic Miner [CITE], the Technology-Topic Framework \#AUTHOR\_TAG, a system that forecasts the propagation of technologies across research communities, and the Pragmatic Ontology Evolution Framework [CITE] & $<$EOF$>$ \\\midrule
\bottomrule
\end{tabular}
\caption{Instance of \dataname\ dataset. First Sentence represents the prefixed sentence, Cited Sentence represents the main cited sentence, and Second Sentence is the suffix sentence. We mark first or second sentence as EOF if the cited sentence is either first or last.}\label{tab:sample_data}
\vspace{-5mm}
\end{table}

\subsection{Dataset}
In this section, we discuss our proposed \dataname\ dataset in detail. Kunnath et al. \cite{kunnath2020overview} introduced the ACT dataset, with annotations for $11,233$ citations annotated by $883$ authors. The cited label was masked with ``\#AUTHOR TAG" denoting the position of the cited object. Additionally, the 3C dataset contained full text and the label denoting the class of a particle citation (c.f. Table \ref{tab:class_count}).

In our work, we extend the 3C dataset to house more discourse information to explain better why a citation is present in a sentence. Our intuition is that the cited sentences mostly either entail or contrast the adjoining sentences. To capture the peripheral sentences, we extract the full-text files corresponding to the COREIDs (unique paper ID) in our dataset to follow through on this discovery. For a given document, we map the location of the main cited sentence and find the prefixed and the suffixed sentence. We use heuristic methods like regex, Levenshtein distance, and hard rules like full-stop identification and author name identification to mark three sentences. In our dataset, the first sentence indicates being the prefix sentence before the citation and second, the suffix sentence after the citation. The six categories of the classes are distributed in labels between  0 and 5, as 0 - BACKGROUND, 1 - COMPARES CONTRASTS, 2 - EXTENSION, 3 - FUTURE, 4 - MOTIVATION, 5 - USES (as suggested in \cite{kunnath2020overview}). Table \ref{tab:sample_data} illustrates a sample instance from \dataname, and
Table \ref{tab:class_count} represents the number of instances per class.


\begin{table}[!t]\centering
\footnotesize
\begin{tabular}{l|r|r|r|r|r|r|r}\hline
\multirow{2}{*}{Dataset} &\multicolumn{6}{c}{Classes} \\\cmidrule{2-7}
&BACKGROUND &COM-CONT &EXTENSN &FUTR &MOTIVN &USES \\\midrule
\dataname\ &1318 &380 &294 &221 &137 &50 \\
\bottomrule
\end{tabular}
\caption{Number of instances in each class. The classes represents Background, COM-CONT: Compare-Contrast, EXTENSN: Extensions, FUTR: Future Works, MOTIVN: Motivation and  Uses.}\label{tab:class_count}
\vspace{-8mm}
\end{table}

\subsection{Implementation Details}
\subsubsection{Dataset Setting:}

On the dataset part, we first preprocess the data by removing stopwords, lowering cases, and removing all special characters. We clip the sentences at 256 token lengths for train and test instances. We use $2400$ instances as a train set and $600$ as a test set.


\begin{table}[!t]\centering

\scalebox{1.0}{
\begin{tabular}{lrrr|rrrr}\toprule
    \multirow{3}{*}{Baseline System} &\multicolumn{6}{c}{Dataset} \\\cmidrule{2-7}
    &\multicolumn{3}{c}{3C} &\multicolumn{3}{c}{\dataname} \\\cmidrule{2-7}
    &Precision &Recall &F1 Score &Precision &Recall &F1 Score \\\cmidrule{1-7}
    Multi-layer Perceptron &0.30 &0.38 &0.30 &0.39 &0.32 &0.39 \\\cmidrule{1-7}
    LSTM-Attention-Scaffold &0.27 &0.48 &0.46 &0.30 &0.55 &0.39 \\\cmidrule{1-7}
    Transformer &0.38 &0.37 &0.34 &0.36 &0.38 &0.37 \\\cmidrule{1-7}
    DistilBERT &0.43 &0.48 &0.37 &0.46 &0.55 &0.40 \\\cmidrule{1-7}
    BART &0.41 &0.46 &0.34 &0.47 &0.52 &0.41 \\\cmidrule{1-7}
    T5 &0.41 &0.43 &0.35 &0.43 &0.48 &0.38 \\\cmidrule{1-7}
    SciBERT &0.45 &0.51 &0.46 &0.52 &0.53 &0.51 \\\cmidrule{1-7}
    \modelname & & & & & & \\
    \modelname\ w/ SF &0.38 &0.37 &0.34 &0.36 &0.38 &0.37 \\
    \modelname\ w/ CTA  &0.34 &0.36 &0.32 &0.38 &0.41 &0.37 \\
    \modelname\ w/ TEA &0.36 &0.43 &0.41 &0.38 &0.42 &0.39 \\
    \modelname\ w/ TEA, SA, CTA &0.46 &0.44 &0.42 &0.60 &0.63 &0.60 \\
    \bottomrule
\end{tabular}}
\caption{Performance benchmarks over the 3C and \dataname\ datasets. We provide six  classification baselines along different ablation versions of \modelname. }\label{tab:result}
\vspace{-8mm}
\end{table}

\section{Baselines}
We discuss the baseline systems in detail. For the language model (LM)-based baselines, we fine-tune LM with the training samples of 3C and \dataname\ till the convergence.
\begin{enumerate}
    \item {\bf Multilayer Perceptron:} We use three stacked dense layers \cite{gardner1998artificial} with softmax activation. We use Glove embeddings as input representation with cross-entropy as loss.
    \item {\bf BiLSTM with Attention and Scaffolding \cite{cohan2019structural}:} This baseline uses BiLSTM with Attention with Glove as input embedding and Elmo as contextual representation. A 20-node MLP is used for the scaffolding task. We preserve all the original hyperparameters for the baseline.
    \item {\bf Transformer:} We use the vanilla Transformer \cite{vaswani2017attention} architecture to run as baseline. We use 4 stacked layers each in Encoder and Decoder with a max sequence length of 32, with softmax as the activation function and cross-entropy as loss.
    \item {\bf DistilBERT \cite{sanh2019distilbert}:} It follows the same architecture of BERT but reduces the number of parameters by making use of knowledge distillation during pretraining. We use the huggingface ported model for the baseline.
    \item {\bf SciBERT:} SciBERT \cite{beltagy2019scibert} uses the standard BERT architecture with pretraining performed on the scientific documents. The hyperparameters are similar to the Transformer baseline.
    \item {\bf BART:} Similar to SciBERT, BART \cite{lewis2019bart} uses a bidirectional encoder along with an autoregressive decoder. It is pre-trained over the Books and Wikipedia data. We use similar hyperparameters to the SciBERT baseline.
    \item {\bf T5:} It  is a pretrained Transformer based encoder-decoder language model \cite{raffel2020exploring}. It is pretrained as a text-to-text Transformer over various supervised and unsupervised downstream tasks. 
\end{enumerate}


\section{Analysis}

We perform ablation studies on our proposed \modelname\ model  to showcase the efficacy of each module. We show that the peripheral context alone can significantly improve the model's performance by providing contextual information. The addition of TEA pushes the performance for each class, concluding that controlled synthetic generation of training data improves the system's overall performance. Table \ref{tab:result} shows that our model attains an improvement of +0.09 F1 points with CTA, SA, and TEA against the best baseline. The improvements are seen in every module. With the introduction of CTA, our model attains an improvement of +0.02 Recall against the base Transformer network. TAdding TEA shows an improvement of +0.02 F1 and +0.04 Recall against the Transformer.

\begin{table}[!htp]\centering
\label{tab:classwise}
\begin{tabular}{llrrrr}\toprule
    Model &Class &Precision &Recall &F1 Score \\\cmidrule{1-5}
    \multirow{6}{*}{\modelname\ } &BACKGROUND &0.67 &0.83 &0.74 \\\cmidrule{2-5}
    &COMPARE CONTRAST &0.49 &0.28 &0.36 \\\cmidrule{2-5}
    &EXTENSION &0.30 &0.09 &0.14 \\\cmidrule{2-5}
    &FUTURE &0.33 &0.17 &0.22 \\\cmidrule{2-5}
    &MOTIVATION &0.55 &0.31 &0.40 \\\cmidrule{2-5}
    &USES &0.61 &0.61 &0.62 \\
    \bottomrule
\end{tabular}
\caption{Class-wise performance metric of the proposed model. We report Precision, Recall and F1 score of each class.}
\vspace{-8mm}
\end{table}

Table \ref{tab:classwise} shows the class-wise performance of \modelname. It shows that our model is able to capture contextual information for all classes. When compared to the best baseline's (SciBERT) confusion matrix in Table \ref{fig:confusionmatrix}, we see that the baseline leans heavily towards the majority class and predicts $0$ for almost all other classes. However, our model was able to predict all classes uniformly. We also analyze the model's prediction errors in Table \ref{tab:errorlabel}. For the sentence in the second row, the model might be distracted by the phrase ``we assess the similarity" giving it the impression that it is a ``use" category. The third row is also likely to be distracted by the phrase ``Following the process of reflection". The mislabelling is probably due to very low number of training samples for these classes. Providing more contextual information and large number of qualitative training samples can help improve the performance of the model.

\begin{table}[!htp]\centering
    \scriptsize
    \begin{tabular}{p{0.85\linewidth}rr}\toprule
    Main Text &True &Pred \\\midrule
    What has been termed episodic foresight (\#AUTHOR\_TAG, 2010), along with autobiographic memory and theory of mind, also makes up much of our mind wandering (Spreng and Grady, 2009), as we preview some future activity or consider possible future options in order to select appropriate action &1 &2 \\ \cmidrule{1-3}
    We assess the similarity of two semantic vectors using the cosine similarity \#AUTHOR\_TAG, since this measure relies on the orientation but not the magnitude of the topic weights in the vector space, allowing us to compare editorial products associated with a different number of chapters &0 &5 \\ \cmidrule{1-3}
    Following the process of reflection presented by \#AUTHOR\_TAG (1996), in the new version, the first word of 4 questions was added as a visual prompt &0 &5 \\ \cmidrule{1-3}
    Some more recent models, though, have also included domain experts to define the learning content of the educational game (\#AUTHOR\_TAG et al., 2017) &4 &1 \\ \cmidrule{1-3}
    \bottomrule
    \end{tabular}
\caption{Sample of \modelname\ classification error on the  \dataname\ dataset.}\label{tab:errorlabel}
\vspace{-8mm}
\end{table}


\begin{table}[t!]
\centering
\scalebox{1.0}{%
\begin{tabular}{l|l|c c c c c c|}
\multicolumn{1}{c}{} & \multicolumn{1}{c}{} & \multicolumn{6}{c}{\bf Predicted} \\ \cline{3-8}
\multicolumn{1}{c}{} & & \multicolumn{1}{c|}{\textbf{Background}} & \multicolumn{1}{c|}{\textbf{Com-Cast}} & \multicolumn{1}{c|}{\textbf{Extensions}} & \multicolumn{1}{c|}{\textbf{Future}} & \multicolumn{1}{c|}{\textbf{Motivation}} & \multicolumn{1}{c|}{\textbf{Uses}} \\ \cline{2-8} 

\multirow{6}{*}{\rotatebox{90}{\bf Actual}} & \textbf{Background}  & \textcolor{blue}{280/320} & 13/0     & 5/0   & 1/0   & 9/0  & 27/0  \\ \cline{2-2} 
& \textbf{Com-Cast}      & 46/73 &  \textcolor{blue}{21/0} & 1/0  & 0/0  & 3/0  & 3/1  \\ \cline{2-2} 
& \textbf{Extensions} & 25/34  & 0/0  &  \textcolor{blue}{3/0} & 1/0  & 2/0  & 3/0 \\ \cline{2-2} 
& \textbf{Future}  & 9/12 & 1/0   & 0/0  &  \textcolor{blue}{2/0} & 0/0  & 0/0 \\ \cline{2-2} 
& \textbf{Motivation}     & 28/55  & 3/0     & 1/0    & 1/0    &  \textcolor{blue}{17/0}     & 5/0 \\ \cline{2-2} 
& \textbf{Uses}  & 30/79 & 5/0   & 0/0   & 1/0 & {0/0} & \textcolor{blue}{59/16} \\ \cline{2-2} \cline{3-8}
\end{tabular}%
}
\caption{Confusion matrix over \modelname\ / SciBERT (best baseline).  The classes represents Background, Com-Cast: Compare-Contrast, Extensions, Future, Motivation and Uses.}
\label{fig:confusionmatrix}
\vspace{-9mm}
\end{table}


\section{Conclusion}
In this paper, we proposed a new dataset \dataname, where we extended the 3C dataset by introducing peripheral contextual information to analyze the contrasting and entailing information. We also introduced a novel model, \modelname\, that uses cross-text attention to attend to the contextual information present in citation input and the peripheral sentences. We also introduced time-evolving augmentation to generate synthetic data for the minority classes during each time step and spatial fusion to attend to the critical information in the input space. Our proposed model achieves a new state-of-the-art on {\dataname} by +0.09 F1 score against the best baseline. We also conducted extensive ablations to analyze the efficacy of the proposed dataset and model fusion methods. For future works, an exciting line of work can be to utilize the discourse information of the sections to provide more context to the inline citations. The contrasting or entailing information in the neighbouring sentence is crucial in understanding the reasoning's of citation intent. Additional tasks like baseline recommendations, scientific paper recommendations, etc., can be greatly improved with the performance improvement over this task.

\section{Acknowledgment}
The research reported in this paper is funded by Crimson AI Pvt. Ltd.

\bibliographystyle{splncs04}
\bibliography{custom}

\end{document}